\documentclass[sigconf, authorversion, nonacm]{acmart}
\usepackage{CJKutf8}

\AtBeginDocument{%
  \providecommand\BibTeX{{%
    \normalfont B\kern-0.5em{\scshape i\kern-0.25em b}\kern-0.8em\TeX}}}

\makeatletter
    
\def\subtasks#1{\g@addto@macro\@keywords{\endgraf\bigskip{\bfseries\Large\noindent SUBTASKS}\endgraf\noindent#1}}%

\def\teamname#1{\g@addto@macro\@keywords{\endgraf\bigskip{\bfseries\Large\noindent TEAM NAME}\endgraf\noindent#1}}%

\makeatother

\begin{document}

\title{A Cascade Model for Argument Mining in Japanese Political Discussions: the QA Lab-PoliInfo-3 Case Study}

\author{Ramon Ruiz-Dolz}

\orcid{0000-0002-3059-8520}
\affiliation{%
  \institution{Valencian Research Institute for Artificial Intelligence}
  \institution{Universitat Politècnica de València}
  \streetaddress{Camino de Vera s/n.}
  \city{València}
  \country{Spain}
  \postcode{46022}
}
\email{raruidol@dsic.upv.es}

\renewcommand{\shortauthors}{Ramon Ruiz-Dolz}

\begin{abstract}
  The rVRAIN team tackled the Budget Argument Mining (BAM) task, consisting of a combination of classification and information retrieval sub-tasks. For the argument classification (AC), the team achieved its best performing results with a five-class BERT-based cascade model complemented with some handcrafted rules. The rules were used to determine if the expression was monetary or not. Then, each monetary expression was classified as a premise or as a conclusion in the first level of the cascade model. Finally, each premise was classified into the three premise classes, and each conclusion into the two conclusion classes. For the information retrieval (i.e., relation ID detection or RID), our best results were achieved by a combination of a BERT-based binary classifier, and the cosine similarity of pairs consisting of the monetary expression and budget dense embeddings.
\end{abstract}

\keywords{Argument Mining, Debate Analysis, Natural Language Processing}

\teamname{rVRAIN}

\subtasks{Budget Argument Mining (BAM)}

\maketitle

\section{Introduction}

The automatic analysis of natural language arguments has made possible to improve computer systems for human assistance in the domains of medicine \cite{mayer2018argument}, academic research \cite{bao2021argument}, web discourse analysis \cite{habernal2017argumentation}, and autonomous debate \cite{slonim2021autonomous} among others. The argument mining task is present in many different domains and instances \cite{lawrence2020argument}. However, due to its heterogeneity and complexity, it is considered as an important challenge in the Natural Language Processing (NLP) research community. The underlying linguistic structures in natural language argumentation present a great challenge for both, the human annotation of new corpora \cite{ruiz2021vivesdebate}; and the training/evaluation of new models for domain-independent argument mining \cite{al2016cross, ruiz2021transformer} or for different instances of the problem belonging to the same domain (e.g., legal) \cite{villata2020using, poudyal2020echr}. Thus, advances in argument mining research will benefit from as many as different viewpoints (e.g., domains and/or task instances) approaching this task.

In this work, we describe the participation of our team \textit{rVRAIN} to the Budget Argument Mining (BAM) task organised for the \textit{QALab PoliInfo 3}\footnote{\url{https://poliinfo3.net/}} and the \textit{NTCIR-16}\footnote{\url{http://research.nii.ac.jp/ntcir/ntcir-16/index.html}}. The BAM is a combination of classification and information retrieval sub-tasks in the domain of political debate analysis. First, the argument classification sub-task is aimed at determining if a given monetary expression belongs to an argument, and which is its argumentative purpose (i.e., either claim or premise). Second, the relation ID detection sub-task is aimed at finding relations between monetary expressions uttered in an argumentative discourse and political budget items.

Our approach presents a BERT-based cascade model for argument mining in japanese political discussions. The model proposed in this paper for solving the BAM task tackles independently the argument classification and the relation ID detection tasks. For the former, we propose the use of handcrafted rules to determine if an expression is monetary or not. Then, a BERT-based cascade model is trained to classify each argumentative monetary expression into premise or claim, and their subsequent sub-classes (i.e., three premise and two claim sub-classes). For the latter, a BERT-based binary classifier is trained to identify possible relations between political budget items and monetary expressions. Each (possible) identified relation is then scored using the cosine similarity, and only the top relations are brought into consideration.

The rest of the paper is structured as follows. Section \ref{relw} reviews the related research and contextualises the contribution of this paper to the area of argument mining. Section \ref{bam} briefly defines the BAM task and the corpus used to carry out our experiments. Section \ref{march} presents the architecture of the model proposed for solving the BAM task. Section \ref{res} depicts the observed results in three different stages of the competition. Finally, Section \ref{disc} discusses the obtained results and analyses future lines of research and open challenges.


\section{Related Work}
\label{relw}

Argument mining approaches the automatic identification, classification and structuring of argumentative natural language \cite{palau2009argumentation}. It has been typically decomposed into different sub-tasks in the literature \cite{lawrence2020argument,bao2021neural}: argumentative discourse segmentation, argument component detection, and argumentative relation identification. Each one tackles a different step belonging to the global goal of argument mining. 

Argumentative discourse segmentation is the task of detecting argument spans in a given natural language input. For example, identifying where an argumentative component begins and ends throughout the full interventions of politicians in a discussion. A basic approach for this task was to consider complete sentences and classify them into \textit{argument}/\textit{non-argument} \cite{palau2009argumentation}. In \cite{levy2014context}, the authors propose an unsupervised approach for claim segmentation based on the appearance of common linguistic structures used for argumentation (e.g., ``that''). However, recent research has emphasised the relevance of context for improving the segmentation of arguments and argumentative components in natural language inputs \cite{ajjour2017unit}. In spoken dialogue it is common to omit contextual information to ease its flow, aimed at overcoming this problem a cascade model for identifying argumentative propositions completing the missing context is proposed in \cite{jo2019cascade}.

The detection and classification of argument components is the argument mining task aimed at understanding the argumentative purpose of the previously segmented text. Research in this topic has usually focused on the identification of argumentative evidence and on the premise/claim classification \cite{lawrence2020argument}. We will focus on the latter since it has a direct relation with the BAM task and the model proposed in this work. The argument component detection is a very descriptive representation of the previously mentioned existing heterogeneity in argument mining research. Initially introduced in \cite{palau2009argumentation}, the task was instanced as a binary classification problem. The authors make use of classical machine learning algorithms to predict premise and claim classes for the argumentative expressions. Subsequent research focused on a linguistic enrichment of the task proposed a new instance where up to four classes (i.e., \textit{major claim}, \textit{claim}, \textit{premise}, \textit{none}) were considered \cite{stab2014identifying}. Some of the latest research in this topic has explored the use of end-to-end neural network-based architectures \cite{eger2017neural,morio2018end}, graph convolutional networks \cite{morio2019syntactic}, and attention-based architectures \cite{stab2018cross} to improve previous experimental results.

Finally, the argumentative relation identification task focuses on detecting argumentative structures between the argumentation components (e.g., premises or claims). This task has been classically considered one of the most complex tasks in argument mining, and approached as a sentence pair classification problem with two classes (i.e., attack and support) \cite{cocarascu2017identifying,eger2017neural,hou2017argument}. Recent research has investigated the behaviour of state-of-the-art NLP techniques when approaching a cross-domain multi-class instance of this task \cite{ruiz2021transformer}. However, since the BAM task and our proposed model does not approach this sub-task of argument mining, we will not go any further into this aspect.

\section{Budget Argument Mining}
\label{bam}

This work approaches the Budget Argument Mining (BAM) instance of the argument mining task. The BAM is aimed at improving the automatic argumentative analysis of political discussion transcripts through the use of NLP techniques. It includes the argumentative discourse segmentation and the argument component detection sub-tasks. For that purpose, monetary expressions are detected in the transcripts, and it must be determined if an expression belongs to an argument or not, and which is its argumentative role in the discussion. Furthermore, the required analysis is enriched with the relation of each monetary expression with a political budget item. This way, the resulting analysis will provide a set of argumentative components and their type detected in the transcripts of a discussion, and a set of relations between the arguments and budget items.

\begin{figure}
    \centering
    \includegraphics[width=0.47\textwidth]{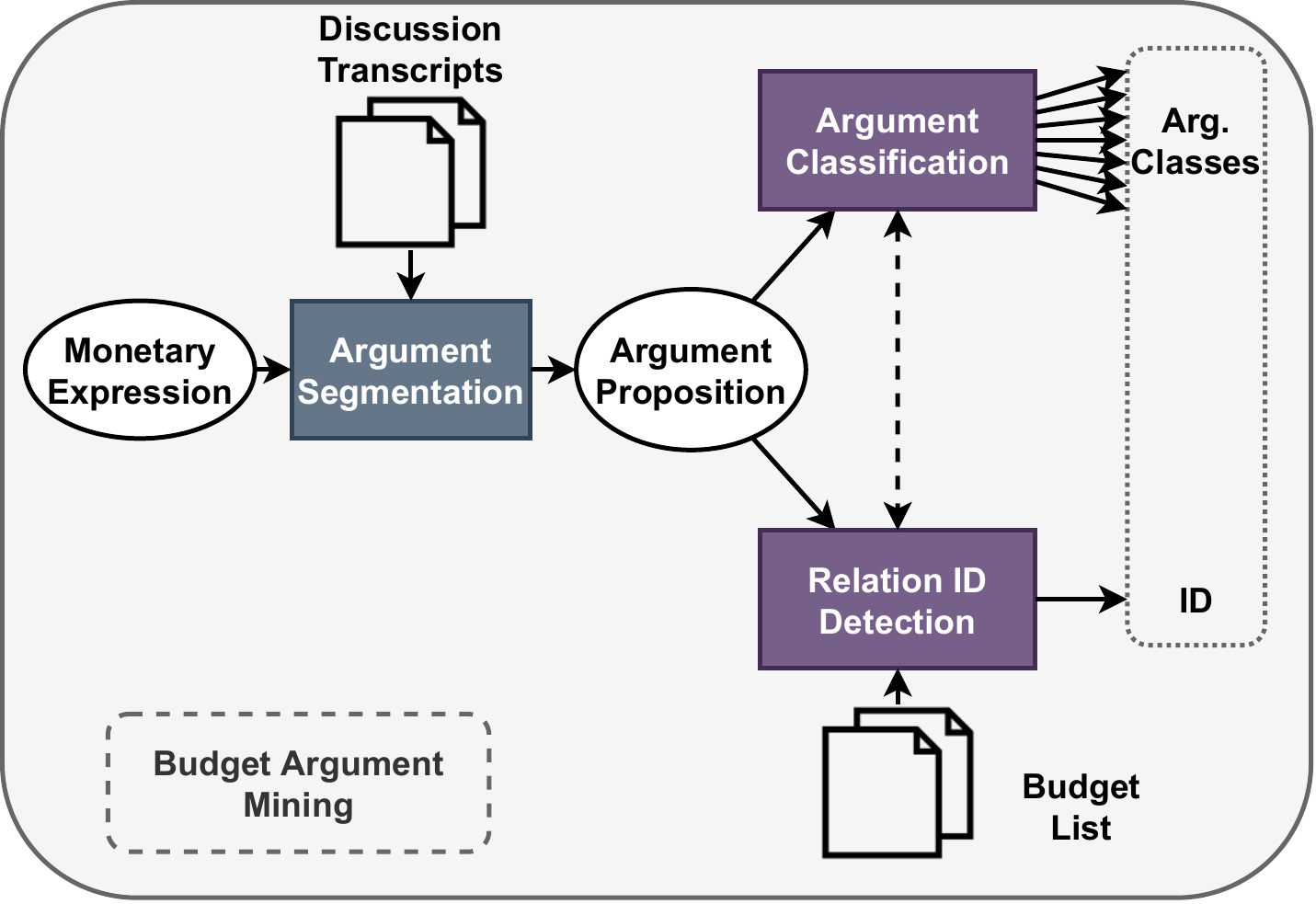}
    \caption{Budget Argument Mining task diagram.}
    \label{fig:bam}
\end{figure}

Therefore, the BAM consists of two different sub-tasks: the argument classification (AC) and the relation ID detection (RID) (see Figure \ref{fig:bam}). A complete description of the whole task can be found in its overview \cite{poliinfo-3-overview}. However, the basic ideas of BAM are presented in the following sections in order to make this paper self-contained.

\subsection{Argument Classification (AC)}

The AC sub-task is aimed at covering the two first parts of argument mining: segmentation and classification. Thus, for a given monetary expression appearing in an utterance, we need to analyse if it belongs to an argumentative proposition, and which is its role in argumentation. First, the argumentative propositions containing the monetary expressions need to be segmented from the natural language transcripts. Second, these segments must be classified into seven different argumentative classes: (i) Premise: Past and Decisions; (ii) Premise: Current and Future; (iii) Premise: Other; (iv) Claim: Opinions, suggestions and questions; (v) Claim: Other; (vi) Not monetary expression; and (vii) Other.

\subsection{Relation ID Detection (RID)}

The RID sub-task is aimed at determining if the monetary expressions uttered in the discussion are related to a specific item in the budget list. For this purpose, each argument containing any monetary expression must be segmented. Then, a relation between the segmented text and the budget items must be established. 

\subsection{Data}

The data released for the BAM task is structured into three different documents: the budget data (\textit{PoliInfo3\_BAM-budget.json}), the training data (\textit{PoliInfo3\_BAM-minutes-training.json}), and the test data (\textit{PoliInfo3\_BAM-minutes-test.json}). Each document contains information from the Japanese national diet, and from three different local circumscriptions (i.e., Otaru, Ibaraki, and Fukuoka).

The budget document is a list consisting of 768 different budget items. Each budget item has eleven descriptive features: an \textit{identifier}, a \textit{title}, a \textit{url}, an \textit{item}, the \textit{budget amount}, a list of \textit{categories}, the \textit{types of account}, the \textit{department}, \textit{last year's budget}, a \textit{description}, and  a \textit{budget difference}. 

The training document contains 29 proceedings belonging to the local circumscriptions. These proceedings consist of a total amount of 1573 utterances. Furthermore, 2 speech records from the national diet consisting of a total of 363 speeches are also included in this file. This translates to a total of 1248 monetary expressions, which are our training samples. The class distribution of these samples is depicted in Table \ref{tab:bamclasses}. The test document follows the same structure. A total of 760 utterances from local circumscriptions and 123 speeches from the national diet are included in this file. From all these transcriptions, 520 monetary expressions remain unlabelled in this document, which is the one used in the model evaluation of the BAM shared task.

\begin{table}
    \centering
    \caption{Class distribution of the BAM training data.}
    \resizebox{0.47\textwidth}{!}{
    \begin{tabular}{l c c c c c c c c}
    \toprule
         & \multicolumn{3}{c}{\textbf{Premise}} & & \multicolumn{2}{c}{\textbf{Claim}} \\ \cline{2-4} \cline{6-7}
         & Past & Future & Other & & Opinions & Other & \textbf{Non monetary} & \textbf{Other} \\ \midrule
        
        \textbf{N} & 260 & 622 & 212 & & 98 & 23 & 27 & 6 \\\bottomrule
    \end{tabular}}
    \label{tab:bamclasses}
\end{table}

\section{Model Architecture}
\label{march}

\begin{figure*}
    \centering
    \includegraphics[width=\textwidth]{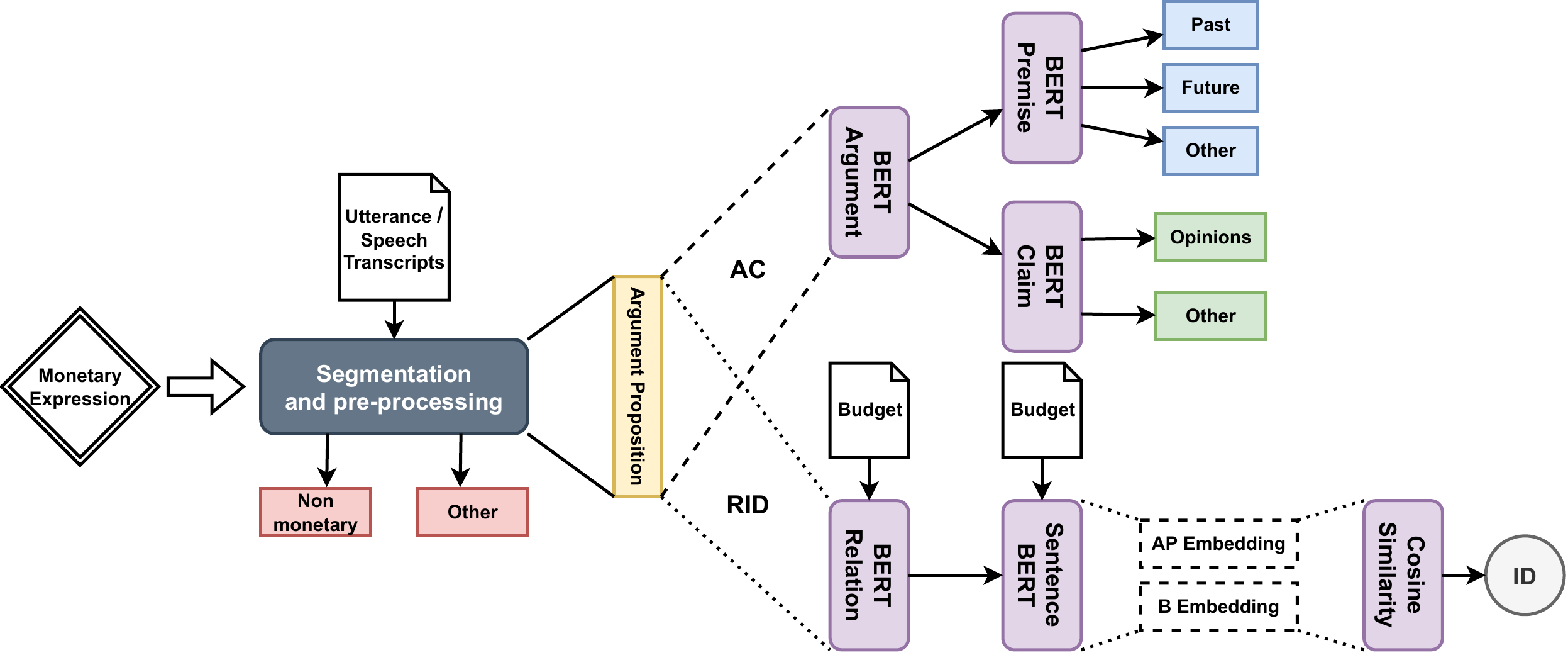}
    \caption{\textit{rVRAIN} model architecture proposal.}
    \label{fig:march}
\end{figure*}

We propose a BERT-based \cite{devlin2018bert} cascade model to undertake the complete BAM process (see Figure \ref{fig:bam}). All the BERT-based classifiers integrated in our cascade model were fine-tuned from the \textit{Inui Laboratory}\footnote{\url{https://github.com/cl-tohoku/bert-japanese/tree/v2.0}} pre-trained BERT-large Japanese Language Model. In our approach, each monetary expression will be treated as the input and a class label and a related ID as the output. The proposed architecture aims to smooth the complexity of the classification task considering the size of the training corpus and the number of classes. Furthermore, it approaches independently the AC and the RID sub-tasks. Figure \ref{fig:march} synthesises the proposed architecture. The code implementation of the model architecture proposed in this paper is publicly available in GitHub\footnote{\url{https://github.com/raruidol/Budget-AM}}.

Before tackling the AC and RID tasks, each monetary expression was analysed together with the discussion transcripts (i.e., local government utterances and national diet speeches) to produce segmented argument propositions. The segmentation was done by considering the set of full sentences belonging to the same utterance/speech where the monetary expression was contained. A set of handcrafted rules was applied during this pre-processing to determine if the proposition was either non monetary or other than a premise or a claim (e.g., detecting the existence of monetary Japanese kanji characters such as ``\begin{CJK}{UTF8}{min}円\end{CJK}''). This way, the total number of remaining classes was reduced from seven to five.

Then, the AC is tackled by three different BERT-based classification models. A high-level BERT-based binary classifier was trained to detect if an argument proposition was either a premise or a claim. Once having assigned a high-level class to the sample, two low-level BERT-based classifiers were trained for 3-class premise classification and binary claim classification. This way, the high-level model focuses on the premise/claim discriminatory features, while the low-level focuses on more specific intra-class features. Furthermore, the class complexity of the problem is also decomposed from 5-class to 3-class and binary classifications.

Finally, the RID is tackled by a BERT-based binary classifier, and a cosine similarity calculation for pairs of Sentence-BERT \cite{reimers2019sentence} embeddings. In this second part of the task, the segmented argument propositions containing monetary expressions are paired together with the \textit{item} and \textit{description} features of the budget items. A binary classifier is used to determine if a given pair (i.e., argument proposition, budget item) could be related or not. Then, all the pairs classified as related are scored using the cosine similarity of the dense embeddings of argument propositions (AP) and budget (B) items (i.e., \textit{item}+\textit{description} features) generated using a Sentence-BERT model. The highest scored relation is used in our approach to produce the model's output.

Therefore, each monetary expression was completely analysed by our cascade model, classified into one of the seven argumentative classes, and related to one of the budget items in the list.

\section{Results}
\label{res}

The evaluation of the architecture proposed in this paper has been carried out at three different levels. First, we performed a local evaluation of the models aimed at having preliminary notions of how would our proposal behave with the test data. Second, we received feedback of our model's performance in an initial ``\textit{Dry-run}'' phase of the BAM task. Finally, the ``\textit{Formal-run}'' evaluation of the models corresponds to the last and definitive round of the shared task.

\subsection{Experimental Setup}

All the experiments and results reported in this paper have been implemented and run under the following setup. For the pre-processing of the corpus, we have used \textit{pandas} \cite{mckinney2010data} for handling data structuring, and \textit{fugashi} \cite{mccann2020fugashi} for the analysis of Japanese natural language text. For model training and transfer learning we have used the \textit{PyTorch} and \textit{Transformers} \cite{wolf2019huggingface} libraries. This powerful deep learning tools have made possible to take advantage of existing large language models in Japanese, and adapt them to our specific task. For the semantic cosine similarity calculus in RID sub-task we used the \textit{Sentence Transformers} \cite{reimers2019sentence} library. Finally, the local evaluation metrics (i.e., accuracy and macro f1) have been implemented using the \textit{sklearn} library.


\subsection{Local Evaluation}

During the local evaluation, we have tested different model architectures. The most basic approach consisted of a 7-class BERT-based classification model (\textit{7BERT}). We also experimented with a 5-class BERT-based classification model together with a set of handcrafted rules for the underrepresented classes (i.e., non monetary and other) (\textit{5BERT}). Finally, we evaluated the BERT-based cascade model proposed in this work for tackling the BAM task (\textit{rVRAIN}). Each of these models was also evaluated considering a balanced version of the corpus, where premise and claim training sample distributions were more balanced than the original corpus (\textit{BD}). For the evaluation, we considered the accuracy and the Macro-F1 scores. This decision was made based on the strong unbalance between classes observed in the training corpus. Furthermore, we evaluated our models using a 10-fold cross validation. Table \ref{tab:loceval} summarises the obtained results during the local evaluation of our models.

\begin{table}
    \centering
    \caption{Local evaluation of the different models for AC.}
    \begin{tabular}{l c c}
    \toprule
        \textbf{Model} & \textbf{Accuracy} & \textbf{Macro-F1} \\ \midrule
        \textit{7BERT} & 0.71 & 0.19 \\ \midrule
        \textit{7BERT(BD)} & 0.56 & 0.16  \\ \midrule
        \textit{5BERT} & \textbf{0.76} & 0.25 \\ \midrule
        \textit{5BERT(BD)} & 0.55 & 0.19 \\ \midrule
        \textit{rVRAIN} & 0.47 & \textbf{0.27} \\ \midrule
        \textit{rVRAIN(BD)} & 0.42 & 0.22 \\ \bottomrule
    \end{tabular}
    \label{tab:loceval}
\end{table}

We can observe how the best accuracy score was obtained by the \textit{5BERT} model. However, \textit{rVRAIN} achieved the best performance considering the Macro-F1 score. This means that our cascade model generalised better on this task, by doing a better classification of the samples belonging to underrepresented classes.

\subsection{Dry-Run Evaluation}

The Dry-Run evaluation phase of the BAM shared task used the test file to evaluate our submissions, and was divided into two different stages. During the early stage (see Table \ref{tab:dry1}), the evaluation script assigned a unique score to the team submissions. This score combined the performance of the models in AC and RID sub-tasks. In the early stage, we evaluated the performance of the same models evaluated during the local evaluation, except for the balanced data versions. Our models achieved the 2nd and 3rd best scores for the BAM task. \textit{RB} stands for the random baseline provided by the organisers of the task.

\begin{table}
    \centering
    \caption{Dry-run (early) evaluation of the different models for BAM.}
    \begin{tabular}{l c}
    \toprule
        \textbf{Team} & \textbf{Score AC+RID} \\ \midrule
        \textit{fuys} & 0.51 \\ \midrule
        \textit{rVRAIN (5BERT)} & 0.45 \\ \midrule
        \textit{rVRAIN} & 0.40 \\ \midrule
        \textit{OUC} & 0.33 \\ \midrule
        \textit{rVRAIN (7BERT)} & 0.25 \\ \midrule
        \textit{RB} & 0.09 \\ \midrule
    \end{tabular}
    \label{tab:dry1}
\end{table}

However, the evaluation script was updated the last month of the Dry-Run evaluation. The late stage (see Table \ref{tab:dry2}) of the Dry-Run evaluation provided individual scores for the AC and RID tasks, together with a global evaluation of the performance of the model in the BAM task. Aimed at easing the readability of the results, we will only include the best performing approaches of each team in the tables. Our best performing model in the late stage was the one using the \textit{5BERT} model for AC, but we could not evaluate the cascade architecture during this phase. Furthermore, the new evaluation script only considered those samples with both, the argument class and the relation ID correctly predicted, to increase the global score of the BAM task. This explains why our approach was the 3rd ranked with the best general score, even though it was the 2nd in the AC and the 1st in the RID sub-tasks.

\begin{table}
    \centering
    \caption{Dry-run (late) evaluation of the different models for BAM.}
    \begin{tabular}{l c c c}
    \toprule
        \textbf{Team} & \textbf{Score AC+RID} & \textbf{AC} & \textbf{RID}\\ \midrule
        \textit{fuys} & \textbf{0.13} & \textbf{0.57} & 0.17 \\ \midrule
        \textit{OUC} & \textbf{0.13} & 0.37 & \textbf{0.21} \\ \midrule
        \textbf{\textit{rVRAIN (5BERT)}} & 0.06 & 0.48 & \textbf{0.21} \\ \midrule
        \textit{takelab} & 0.00 & 0.33 & 0.00 \\ \midrule
        \textit{RB} & 0.00 & 0.13 & 0.00 \\ \midrule
    \end{tabular}
    \label{tab:dry2}
\end{table}

\subsection{Formal-Run Evaluation}

During the Formal-Run evaluation, the same test file than with the Dry-Run was used. We achieved our best results using the proposed cascade model architecture for AC, together with the proposed semantic similarity calculation method for RID. As presented in Table \ref{tab:formal}, the \textit{rVRAIN} achieved the 4th best performing position from a total of 6 participating teams. However, our approach was the best performing one from the teams that did not include task organisers.

\begin{table}
    \centering
    \caption{Formal-run evaluation of the different models for BAM. (*)The team contains task organisers.}
    \begin{tabular}{l c c c}
    \toprule
        \textbf{Team} & \textbf{Score AC+RID} & \textbf{AC} & \textbf{RID}\\ \midrule
        \textit{JRIRD}* & \textbf{0.51} & \textbf{0.58} & 0.61 \\ \midrule
        \textit{OUC}* & 0.45 & 0.57 & \textbf{0.66} \\ \midrule
        \textit{fuys}* & 0.23 & 0.57 & 0.34 \\ \midrule
        \textbf{\textit{rVRAIN}} & 0.17 & 0.48 & 0.21 \\ \midrule
        \textit{rVRAIN (5BERT)} & 0.06 & 0.48 & 0.21 \\ \midrule
        \textit{takelab} & 0.04 & 0.39 & 0.06 \\ \midrule
        \textit{SMLAB} & 0.00 & 0.38 & 0.00 \\ \midrule
        \textit{RB} & 0.00 & 0.13 & 0.00 \\ \midrule
    \end{tabular}
    \label{tab:formal}
\end{table}

\section{Discussion}
\label{disc}

We have described the participation of \textit{rVRAIN}'s team at the \textit{Budget Argument Mining} task organised in the \textit{QALab PoliInfo 3} and the \textit{NTCIR-16}. The organisers proposed a new instance of the argument mining task, a classic in the NLP area of research. In this new instance, the main goal was to correctly classify arguments containing monetary expressions and relate them to items in a list of political budgets. In this paper, we have proposed a new approach to this task relying in the latest advances in NLP (i.e., Transformer-based architectures). The proposed cascade model architecture achieved the fourth position in the performance ranking, and it was the best among teams without task organisers.

Several observations can be drawn from this paper's proposal and experimentation. First, we have seen how when dealing with highly unbalanced corpora, a system can benefit from defining a set of handcrafted rules and relaxing the class complexity of the task. Instead of approaching the complete problem with the use of a unique classifier. Second, we have also observed that no improvement could be achieved by forcing the balance of the corpus. When using the balanced version, the score dropped significantly. This is most probably because the real distribution that the model has to predict is not balanced, but the corpus size limitation can also have a major role in this issue.

Finally, we foresee the implementation of some communication between the models for AC and RID during their training as a future work improvement of the model's performance on the BAM task. Furthermore, we also consider that using different test sets for each phase of the shared task (i.e., Dry-Run and Formal-Run) would be beneficial for the generalisation of the findings in this topic.

\begin{acks}
This work is supported by the Spanish Government project PID2020-113416RB-I00.
\end{acks}

\bibliographystyle{ACM-Reference-Format}
\bibliography{biblio}

\end{document}